**Title:** Bio-Inspired Compensatory Strategies for Damage to Flapping Robotic Propulsors

**Authors:**

M. L. Hooper,[1]* I. Scherl,[2] M. Gharib[1]

**Affiliations:**

[1]Graduate Aerospace Laboratories, Division of Engineering and Applied Science, California Institute of Technology; Pasadena, CA, USA.

[2]Department of Mechanical and Civil Engineering, Division of Engineering and Applied Science, California Institute of Technology; Pasadena, CA, USA.

*Corresponding author. Email: mlhooper@caltech.edu.

**Abstract:** To maintain full autonomy, autonomous robotic systems must have the ability to self-repair. Self-repairing via compensatory mechanisms appears in nature: for example, some fish can lose even 76% of their propulsive surface without loss of thrust by altering stroke mechanics. However, direct transference of these alterations from an organism to a robotic flapping propulsor may not be optimal due to irrelevant evolutionary pressures. We instead seek to determine what alterations to stroke mechanics are optimal for a damaged robotic system via artificial evolution. To determine whether natural and machine-learned optima differ, we employ a cyber-physical system using a Covariance Matrix Adaptation Evolutionary Strategy to seek the most efficient trajectory for a given force. We implement an online optimization with hardware-in-the-loop, performing experimental function evaluations with an actuated flexible flat plate. To recoup thrust production following partial amputation, the most efficient learned strategy was to increase amplitude, increase frequency, increase the amplitude of angle of attack, and phase shift the angle of attack by approximately 110 degrees. In fish, only an amplitude increase is reported by majority in the literature. To recoup side-force production, a more challenging optimization landscape is encountered. Nesting of optimal angle of attack traces is found in the resultant-based reference frame, but no clear trend in amplitude or frequency are exhibited-- in contrast to the increase in frequency reported in insect literature. These results suggest that how mechanical flapping propulsors most efficiently adjust to damage of a flapping propulsor may not align with natural swimmers and flyers.

## INTRODUCTION

One important challenge faced by autonomous robotic systems is robustness. To remain fully autonomous, systems must have the capability to self-repair after damage. Even if we were to accept a reduction in level of autonomy during the repair process, many long-duration remote deployments preclude access to non-autonomous repairs. Thus, we must seek a solution that will allow the autonomous robotic system to repair itself.

The ability to self-repair upon damage appears in nature in many forms. As one example, the brittle star *Amphiura filiformis* can fully regrow an amputated appendage (*1*). However, this fully self-regenerative process is far from implementable in robotic systems currently. We must instead turn to alternative methods for self-repair upon damage. Many fish and insects, for whom self-regeneration is either slow or impossible, can repair functionality by altering stroke mechanics after damage (see Table S1). We seek to determine how robotic systems may similarly self-repair through compensatory behaviors after damage.

We choose flapping propulsion as our candidate for investigation for two main reasons. First, nature already shows a remarkable ability to overcome damage in this mode. Some animals that employ flapping production may lose even 76 percent of their propulsive surface without significant disruption in thrust production, due to compensatory alterations in stroke mechanics (*2*). Second, flapping propulsion is implementable on a variety of autonomous robotic systems including both autonomous underwater vehicles (AUVs) and micro-air vehicles (MAVs). Determining a strategy to self-repair after damage to flapping propulsion would therefore be widely applicable.

When natural swimmers and flyers experience catastrophic propulsor damage, certain compensatory behaviors enable them to recoup force production. These behaviors have been well-studied in a variety of animals, including bees (*3-6*), moths (*7,8*), butterflies (*7,9*), fruit flies (*10*), cyprinid fish (*2,11, 12*), mosquitofish (*13*), salmon (*14*), whiting (*15*), and brown darters (*16*, for a summary of compensatory behaviors see Table S1). For example, in (*13*) the relationship between mosquitofish swimming performance and caudal fin size was observed to follow a Gaussian curve, where critical sustained swimming speed was maintained with a moderate amount of damage, but both larger damaged fins and smaller damaged fins experienced a loss. Specific compensatory behaviors such as amplitude or frequency alteration enabling maintenance of critical sustained swimming speed with a moderate amount of damage were not recorded in (*13*). However, these adjustments may be driven by selective pressures, anatomical constraints, or other factors that are irrelevant to robotic propulsors. In fact, the mosquitofish is known to display intraspecific aggression and likely has excess caudal fin area to buffer against aggression-related fin damage (*13*). A robotic system is not subject to the same selective pressure (i.e., intraspecific aggression) and therefore may not require similar alterations (i.e. increasing fin area) to buffer against propulsor damage. It is not yet known whether bio-inspired compensatory behaviors like those exhibited by natural swimmers and flyers are the most efficient strategy to mitigate damage in robotic flapping propulsors. Much previous work has focused on faithfully reproducing biological behaviors on robotic systems to provide more experimental accessibility (*17, 18*). Instead, we explore what methods of adapting to propulsor damage are most efficient in a robotic system.

To determine which compensatory mechanisms are most efficient, we take the bio-inspired approach of artificial evolution. Artificial evolution has been applied successfully to various problems including cylinder drag (*19*), flapping plates (*20-23*), aerodynamic shape optimization (*24*, reviewed in *25),* and more (*26-28*). For our artificial evolution implementation, we use the structure of (*22*): a Covariance Matrix Adaptation Evolutionary Strategy (CMA-ES) developed by (*29-32*) in the loop with experimental function evaluations.

In this work we seek to determine which compensatory mechanisms allow a robotic flapping system to compensate for catastrophic damage most efficiently. We equip a robotic flapping propulsor with an evolutionary machine learning algorithm (CMA-ES), utilizing a hardware-in-the-loop architecture. The robotic system can actuate trajectories with more extreme flapping angles than biological systems, leading to an increased search space for the machine learning algorithm. During the optimization, we intentionally damage the propulsor by amputating approximately 50 percent of its area. The optimal compensatory behaviors determined via efficiency optimization were compared to existing compensatory behaviors by natural swimmers and flyers and found to differ in several ways. This difference suggests that the most efficient adjustments to damage of a robotic propulsor may not mimic all adjustments shown by natural swimmers and flyers.

## MATERIALS AND METHODS

This work utilizes the experimental setup and trajectory generation of (*22*) with modifications to fin size, shape, and flexibility. A 200 x 50 mm flexible fin made of 0.02" thick Delrin ($\rho$ = 1420 kg/m$^3$, E = 3.3 GPa) is attached to the end of a 3D-printed polylactic acid (PLA, $\rho$ = 1300 kg/m$^3$, E = 3.5 GPa) rod. The rod is actuated by a spherical parallel manipulator (SPM) designed by (*33, 34*), moving the fin in a user-defined trajectory (Figure 1.A). The SPM can achieve both infinite rotation about the axis of the rod and large-angle deflections along both x- and y-axes of the lab frame, with accuracy within 1°. The fin is submerged within a 1.0 x 2.4 x 1.2m tank containing Chevron Superla white oil ($\rho$ = 880 kg/m$^3$, $\nu$ = 115 cSt). Oil, as opposed to water or air, is used in this study to increase the signal-to-noise ratio, and results in a range of Reynolds number based off fin span and average velocity per cycle of approximately Re = 440 - 960 for all optima found in experiment. For comparison in the realm of thrust production, *Spinibarbus sinensis, Cyprinius carpio,* and *Carassius auratus* have Reynolds numbers of 535, 633, and 1264 when estimated using measurements from (*2*). For comparison in the realm of side force production, the Reynolds number of *Drosophila hydei* was estimated to be 506 using measurements from (*10*) and (*35*). Note that wing measurements used in our estimate were scaled by the body length ratio of *D. melanogaster* to *D. hydei*.

While moving through the working fluid, the mounted fin experiences forces and moments that are measured by a six-axis ATI Nano-25 Force/Torque sensor. The rod extends 225mm to avoid any possible surface effects, and the SPM is mounted equidistant from three walls of the tank to minimize wall effects.

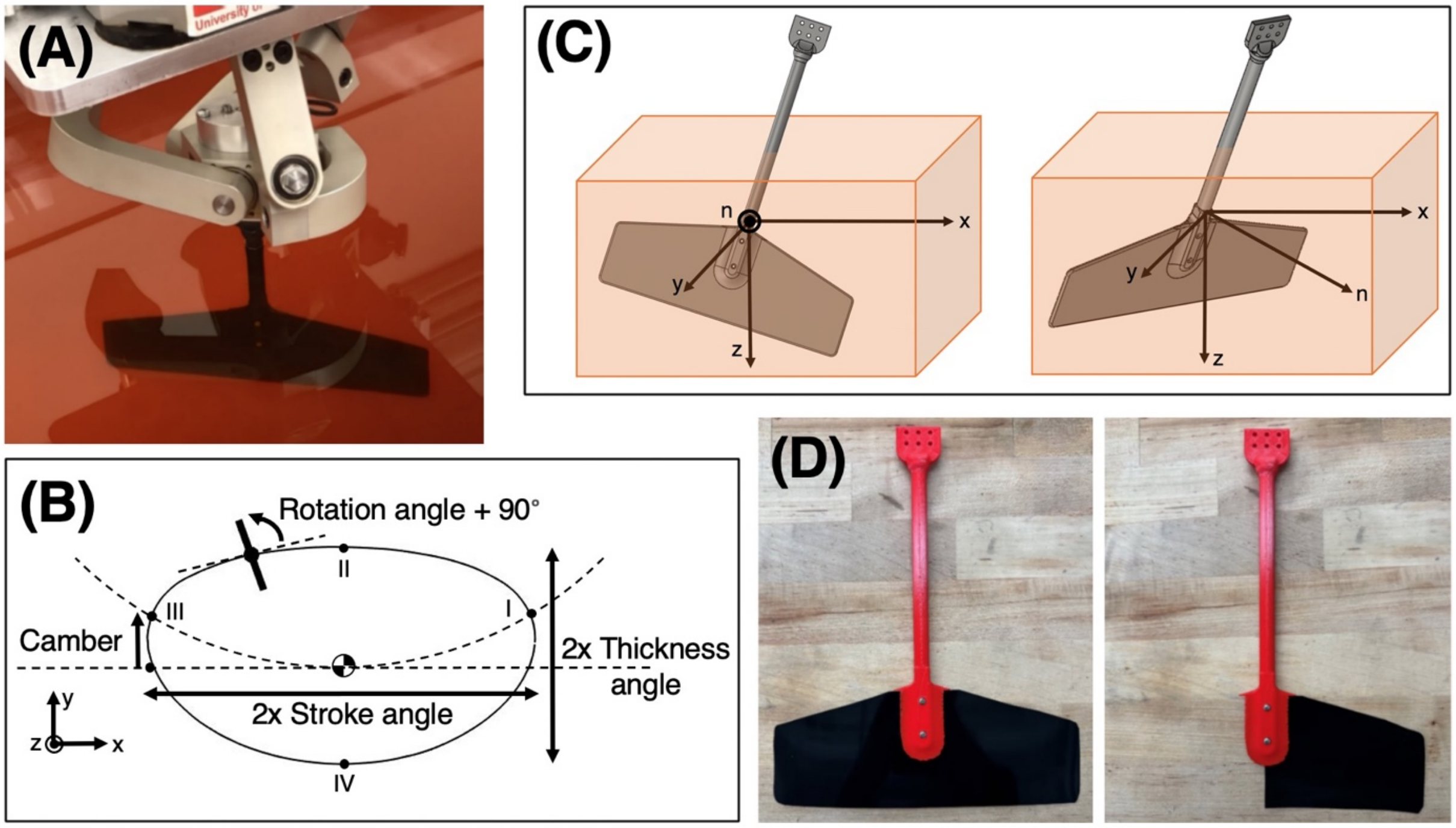

**Fig. 1. Views of the experimental setup and parameters.** (**A**) shows the fin attached to the SPM and submerged inside the oil tank. (**B**) is a visual representation of selected trajectory parameters. The trajectory is viewed from the underside of the fin (i.e., positive z out of the page). The fin, shown by a solid line intersecting the elliptical trajectory, moves counterclockwise passing through points I, II, III, and IV in order. (**C**) defines the coordinate system used in this work; the x-, y-, and z-coordinates are all defined in the lab frame. The normal coordinate, denoted "n", is normal to the plane defined by a fully rigid fin. The two types of flexible fins, intact and amputated, are shown in (**D**).

The motion of the SPM is parametrized by the ten variables listed in Table 1. For this work, we fix the type of trajectory to an ellipse, allowing nine parameters to participate in the optimization. Selected parameters are visualized in Figure 1.B. Three parameters, the stroke angle, thickness angle, and camber, define the geometry of the trajectory. The stroke angle is the angle required to be actuated by the SPM at the furthest x-excursion of the trajectory; in other words, it is a parametrization of the semimajor axis of the ellipse. The thickness angle is the angle required to be actuated by the SPM at the furthest y-excursion of the trajectory; in other words, it is a parametrization of the semiminor axis of the ellipse. The camber defines the degree of curvature of the major axis of the ellipse. Three parameters define the angle of the fin with respect to the trajectory: the rotation angle, rotation phase, and rotation acceleration. The rotation angle is the maximum angle which the fin makes along the trajectory with a line perpendicular to the local instantaneous slope of the ellipse. The rotation phase defines at which azimuthal position that maximum fin angle occurs. The rotation acceleration parametrizes how quickly the fin angle changes from its maximum to minimum values. The higher the value of the rotation acceleration, the closer to a square wave the fin's angular trace appears. Lastly, three parameters define the speed at which the trajectory is actuated. The trajectory is not all actuated at a constant velocity-- half of the trajectory experiences a relative speed-up. This half of the trajectory is defined by the floor of the speed code. A speed code of 1 corresponds to the section I-II-III, 2 to II-III-IV, 3 to III-IV-I, and 4 to IV-I-II. The relative speed-up of this section is dictated by the speed-up value, where for example a value of 1.1 would correspond to a 10% increase in speed. The frequency dictates the average speed at which a full cycle of the trajectory is executed. For further detail on the calculations required to convert these parameters to time-resolved positions of the fin, please refer to (*22*).

| | **Minimum** | **Maximum** | **Convergence criterion** | **Thrust initialization** | **Side force initialization** |
|---|---|---|---|---|---|
| **Type** | Ellipse (0) | Figure Eight (1) | 0.5 | Ellipse (0) | Ellipse (0) |
| **Stroke angle** | 15.27° | 32.18° | 3° | 25.43° | 26.12° |
| **Thickness angle** | 0° | 15.27° | 3° | 14.29° | 13.30° |
| **Rotation angle** | -70° | 70° | 3° | -40.46° | -62.51° |
| **Rotation phase** | 0 | $2\pi$-0.1 | 0.4 | 6.18 | 2.63 |
| **Speed code** | 0 | 4.9 | 0.9 | 2.95 | 1.34 |
| **Speed-up value** | 1.1 | 1.3 | 0.1 | 1.10 | 1.18 |
| **Rotation acceleration** | 0 | 1 | 0.2 | 0.13 | 0.03 |
| **Camber** | 0 | 1 | 0.2 | 0.20 | 0.18 |
| **Frequency** | 0.7 | 0.9 | 0.01 | 0.71 | 0.72 |

**Table 1. Trajectory parameters' range, convergence criteria, and initializations.** The minimum and maximum values possible for each trajectory parameter are reported, along with their convergence threshold and initialization values for both thrust and side force trials. Initialization values are those that were found optimum for an inflexible PLA fin of equal aspect ratio and size, using the same fitness function.

The experimental apparatus and trajectory parametrization described above is paired with a Covariance Matrix Adaptation Evolutionary Strategy (CMA-ES) as the learning algorithm for this work (*30, 32*). We initialize CMA-ES with trajectory parameter values that we found optimal from a previous optimization run with the same fitness function for an inflexible PLA fin of the same aspect ratio and size (Table 1). Here we provide a simplified process overview of the CMA-ES applied to our system. First, a set of candidate solutions (trajectories) are generated from a multivariate normal distribution. The individual real-world fitnesses of this group of trajectories, called a "generation", are assessed using the experimental setup. Every ten generations of an optimization takes approximately 13 hours to complete, due to the in-the-loop evaluations of the fitness function. Each trajectory is actuated at least three times to achieve consistent results. Furthermore, to avoid start-up effects, the first three cycles of each trajectory are removed from the data. The experimentally determined fitness values of each trajectory in the current generation determine the multivariate normal distribution from which the next generation of trajectories is drawn, since these fitness values alter the mean and covariance matrix that define the probability distribution. Once a new generation has been generated from the updated probability distribution, the experimental evaluations begin again. Generations are created and evaluated until the algorithm reaches all convergence criteria. The convergence criterion for each trajectory parameter is reported in Table 1. As per (*22*), the best performing trajectory in the final generation is taken as the optimum, since the best-performing trajectory over all generations is sensitive to noise.

CMA-ES attempts to find the global minimum of the fitness function it is given. In this work, the fitness of each trajectory is calculated by Equation 1, where F is either the force in the z-direction (if optimizing for thrust) or the magnitude of the resultant force in the x-y plane (if optimizing for side-force). For these experiments, a target force $F_{target}$ of 1N was defined.

$$f = 0.8\frac{\left|F_{target} - |F|\right|}{F_{target}} + 0.2\left|1 - \left|\frac{F}{F_n}\right|\right| \quad (1)$$

The first term in Equation 1 enforces the closeness of the produced force (F) to the target force ($F_{target}$) while the second term is a measure of geometric efficiency. Geometric efficiency quantifies the portion of the force produced normal to the fin ($F_n$) that is in the desired direction of force production. The coordinate normal to the plane defined by a fully rigid fin, labeled "n", is visualized in Figure 1.C. The prefactors on each term, 0.8 and 0.2 respectively, were chosen by (*22*) and verified via a sensitivity analysis. Therefore, this fitness function will optimize for a trajectory that produces the desired amount of force in the desired direction, with high geometric efficiency.

In both our experiments optimizing for side-force and for thrust, we introduce fin breakage during the optimization. An optimization is initiated with a full fin (Figure 1.D).

The initial parameters for this optimization are set by the optimal parameters found for an inflexible fin of equal aspect ratio and size via the same fitness function. After actuation and evaluation of 69 generations, the optimization state is copied. The full fin is then allowed to finish optimization, which in the case of thrust optimization occurs ten generations later, and for side-force optimization occurs eleven generations later. This produces two optimization states; one, the copied full-fin state that is close to convergence and two, the optimized full-fin state. To compare the optimal trajectory for an unbroken fin to the learning algorithm's adaptation to a broken fin, we then resume multiple optimizations at Generation 70 from the copied full-fin state. Each of these resumed optimizations now uses a fin where the left half has been removed (Figure 1.D). The half that remains is 180 degrees from the defined angle of attack, as measured azimuthally. The mounting point of the rod to the fin is not altered, and thus removal of half the fin results in a total reduction in propulsive area of 44.2%.

# RESULTS

## Thrust Production

### *Fitness and Force Production*

Immediately upon amputation of half the fin, the CMA-ES algorithm is exploring a parameter subspace that it has found to be optimal for the whole fin. However, experimental evaluations are now performed with an amputated fin. This mismatch, where trajectories that are in the optimal basin for the whole fin are evaluated on the partial fin, causes a distinct increase in the fitness function and distinct decrease in force produced. A higher fitness value signals worse performance (Equation 1). These sharp changes in both fitness and force are easily identified in the top two rows of Figure 2.A and corroborate the recorded generation at which the fin was modified during optimization. After amputation, the CMA-ES algorithm continues to iteratively alter its search space, narrowing in on a new optimal basin for the amputated half fin.

Remarkably, the fitness and force production of the amputated propulsors optimizing for thrust recover fully by the time of convergence, despite a propulsive area loss of almost 50% (Figure 2.A). The fitnesses of the amputated fins are all lower than the fitness of the intact fin, and the closeness to setpoint force varies from the intact fin by at most 1.01%. Although the majority of natural swimmers in the literature are not able to maintain their thrust production after catastrophic loss of propulsor area (see Table S1), many were subject to removal of 100% of their propulsive area. Of those fish for whom approximately 50% was removed (*Mylopharyngodon piceus, Etheostoma edwini, Spinibarbus sinensis, and Cyprinius carpio*), an equal split is observed between maintenance or decrease of thrust production in response to caudal fin damage. While natural swimmers, depending on the individual species, may or may not maintain efficient force production following a 50% reduction in caudal fin area, our learned approach recovers from such damage.

The force traces throughout the converged cycle further corroborate that force production has returned to its predamaged baseline. The average forces for all trials, including the intact fin, differ from the setpoint of 1N by 1.3% or less. Qualitatively, all force traces are similar (Figure 2.B) although some phase shifts are seen in Amputated 1 and 3. These can be quantified by a Fourier decomposition (Table S3). We also see an elongated bilobed shape appear in the force traces, where minima occur at approximately 90 and 300 azimuthal degrees and maxima at approximately 0 and 180 azimuthal degrees. This bilobed shape arises from the mechanics of the spherical parallel manipulator. Since the fin stem remains a constant radius and all force is produced normal to the fin planform, the component of normal force projected onto the z-direction is thereby largest at 0 and 180 degrees.

***Trajectory Parameters: Amplitude, Frequency, and Angle of Attack***

To maintain average force production with a smaller propulsive area, it has been posited in theory (*36*) that both amplitude and frequency must increase. Indeed, the majority of in vivo studies of fish have found an increase in amplitude after catastrophic damage. However, in vivo studies of fish have not shown a consensus on frequency alteration after fin damage (*2, 11-16,* Table S1).

In our experiments, the optimization increases both the converged amplitude and frequency after breakage to maintain efficient thrust production, matching theory (*36*), but not in vivo experiments. The converged amplitudes for the amputated fin are higher than the unbroken optimal amplitude by between 5.2 and 7.2 degrees (Figure 2.A). This difference is significant, as it is larger than the convergence criterion of 3 degrees. Most of the amputated trials attained a higher converged frequency than that of the intact fin, ranging from 0.02 to 0.08 Hz greater. This difference is larger than the convergence criterion of 0.01 Hz. However, it should be emphasized that one trial did not display an increase in the optimal frequency of motion (notably, Optimization 2; see Table S2).

The angle of attack throughout the cycle is also significantly changed by the optimization as a response to damage, leading to an increase in the maximum angle of attack and an approximately 110-degree phase shift (Figure 2.B). A paired visualization of the amplitude and angle of attack can be seen in Figure 2.C. While the increase in maximum angle of attack is a clear mechanism to provide higher forces, the large phase shift is intriguing. This stark difference between the intact fin's optimal angle of attack over a cycle and the amputated fins' has not been reported in the literature on in vivo amputation studies. Thus, this may be a distinct difference between the biological and machine-learned adaptations to produce equal thrust despite propulsor damage.

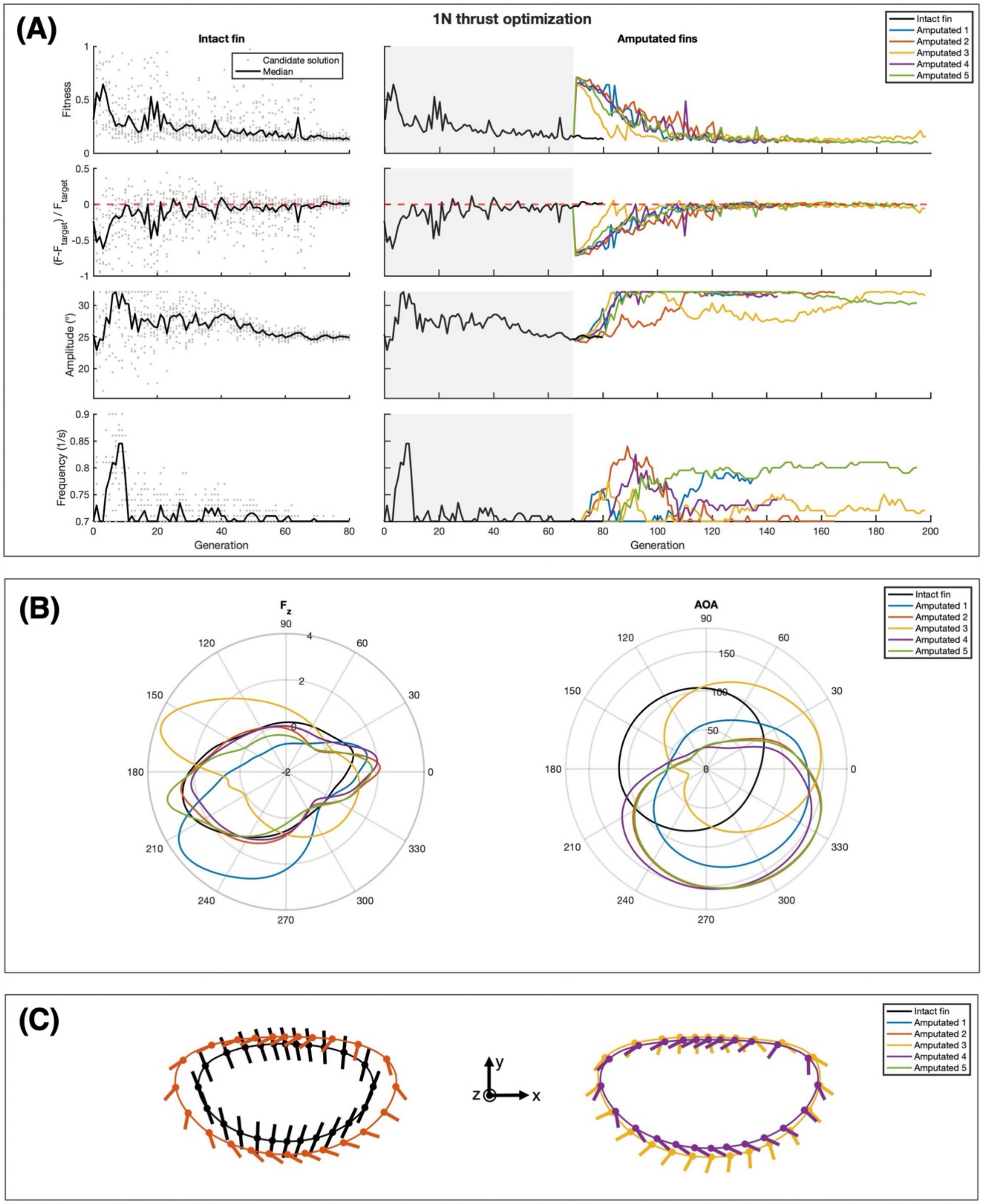

**Fig. 2. Optimization for 1N of thrust force.** In subfigure (**A**), the column titled "Intact Fin" shows the optimization path for the intact fin optimizing for 1N of thrust. The median values for each generation are connected by a line, while experimentally determined fitness values for each individual candidate solution are represented by dots. The column titled "Amputated Fins" shows the progression of the learning algorithm for all fins, omitting individual candidate points for clarity. Note that Generations 0-69, shaded in gray, are identical to the intact fin. The optimization was resumed with each of the amputated fins from Generation 70 of the intact fin. (**B**) shows polar plots of the force production and angle of attack of each trajectory. Angle of attack is measured azimuthally. (**C**) shows the optimal trajectories for thrust production, projected onto the x-y plane. The fin progresses counterclockwise, and the angle of attack of the fin stem is plotted every 15 azimuthal degrees. The left shows the intact fin trajectory compared with the median amputated fin solution (Amputated 2). The right shows the two most different amputated fin solutions (Amputated 3 and 5), whose spatial bounds encompass all other amputated fin solutions.

**Side Force**

### *Fitness and Force Production*

When we repeat the same experimental procedure and instead optimize for a force in the x-y plane, here called side force, we no longer see as straightforward a recovery. The fitness and force production of the side force trials are worse in all cases and are notably not able to fully recover in Amputated 1. In contrast to the thrust trials, where all amputated fins displayed lower fitnesses than the intact fin, all amputated fins in the side force trials displayed higher fitness than the intact fin by 5-19 times. Their closeness to setpoint force were also highly impacted. All amputated trials except for Amputated 1 fall within two percent of the desired target force- a similar range to the thrust amputated trials. However, more of the side force trials exist near to that 2% maximum. Finally, Amputated 1 is clearly unable to recover its predamaged force production, as it converges to a solution with a 17% deficit in thrust.

To analyze the force production of the side force cases, we must first note that the desired 1N of side force was allowed to take any angle within the x-y plane during the optimization. For this reason, Figure 3.B displays plots where the resultant forces are aligned (i.e., the desired 1N of side force is along the x*-axis). Figure 3.C shows the relationship between trajectories in the original x-y reference frame and the trajectories in the x*-y* reference frame. In the x*-y* reference frame, the force and angle of attack profiles collapse well (Figure 3.B).

### *Trajectory Parameters: Amplitude, Frequency, and Angle of Attack*

The amplitude of motion increases when the fin is optimizing for side-force production and spontaneous amputation occurs, though to varying degrees. As seen in Figure 3.A, Amputated 1, 2, and 3 display significantly larger converged amplitudes than the intact fin (32.1, 31.3, and 29.6 degrees versus 15.2 degrees), while Amputated 4 and 5 show a lower increase (18.7 and 16.8 degrees, respectively). Given that the optimal basin has a 3-degree amplitude tolerance required for convergence, we can state that Amputated 5's amplitude is within the optimal basin of the intact fin, while Amputated 4's amplitude is barely larger. We thus find that the first three amputated trials support a significant increase in amplitude to recover from a spontaneous amputation, while the last two trials do not. This learned data is consistent with the approximately equal split of data from in vivo experiments; for those species in which amplitude was reported, it either increased or did not change.

The effect on frequency is even less distinguishable; two out of five amputated trials achieve a higher frequency than the optimal intact fin, one achieves the same frequency, and two out of five attain a lower frequency. However, the in vivo literature shows by a large majority that frequency tends to increase when compensating for propulsor damage.

The angle of attack has a similar phase between the intact fin and the amputated fins when the trajectories are rotated into a common force-defined reference frame (Figure 3.B). This similarity stands in contrast to the production of thrust (Figure 2.B), where the dominant Fourier mode of the intact solution displayed a 110.6-degree phase shift from the average amplitude of the amputated solutions' dominant Fourier mode (Table S3). Furthermore, Figure 3.B shows the nesting of all angle of attack optima. For all trials, whether intact or amputated, the angle of attack traces form concentric shapes. The traces are nested following the order of resultant forces in Figure 3.C. Progressing clockwise from the resultant for Amputated 1 is equivalent to progressing outwards in the nesting order. This implies that a simple strategy for controlling the direction of the force produced in the x-y plane would be to add or subtract angle of attack throughout the cycle.

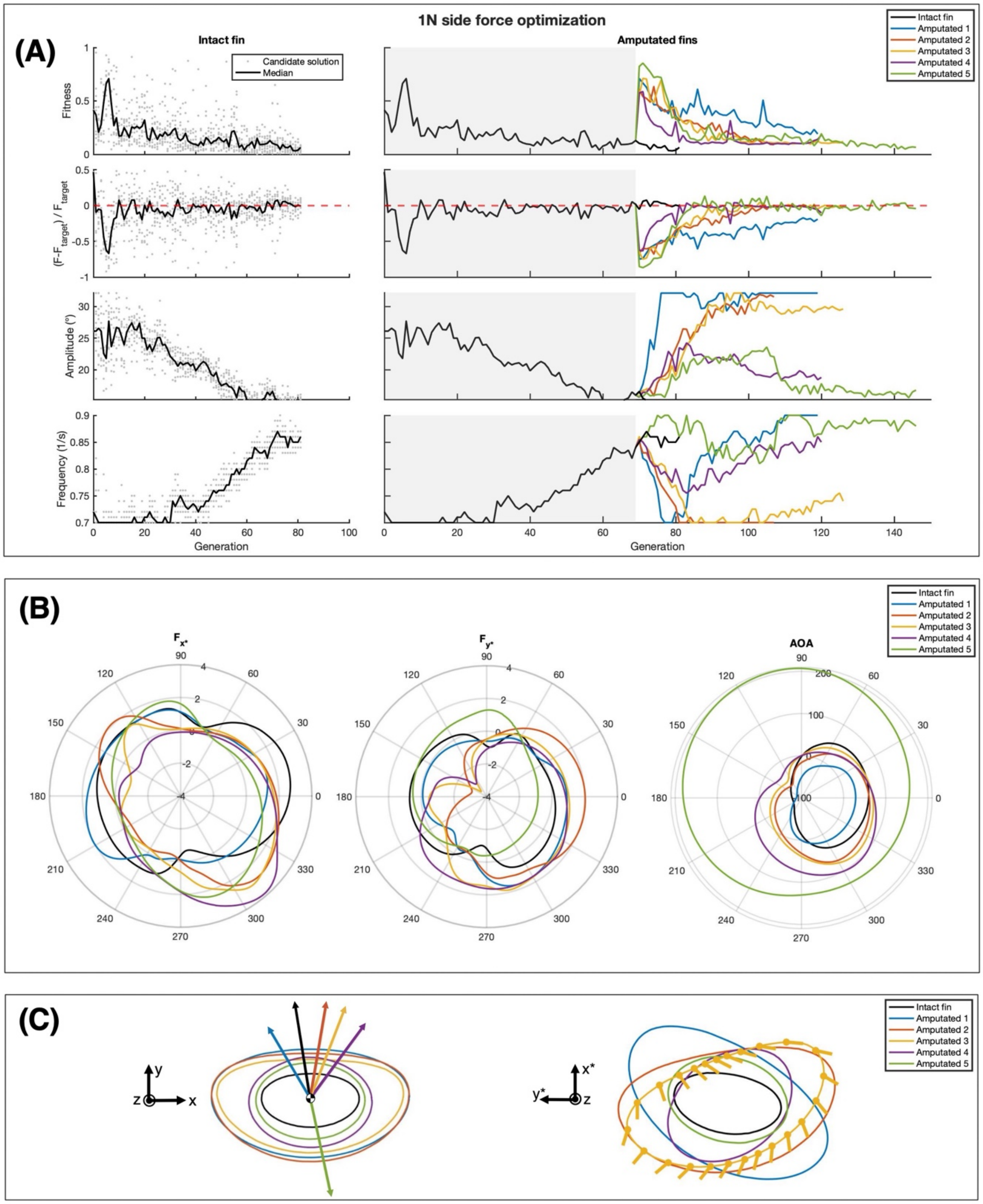

**Fig. 3. Optimization for 1N of side force.** In subfigure (**A**), the column titled "Intact Fin" shows the optimization path for the intact fin optimizing for 1N of side force. The median values for each generation are connected by a line, while experimentally determined fitness values for each individual candidate solution are represented by dots. The column titled "Amputated Fins" shows the progression of the learning algorithm for all fins, omitting individual candidate points for clarity. Note that Generations 0-69, shaded in gray, are identical to the intact fin. The optimization was resumed with each of the amputated fins from Generation 70 of the intact fin. (**B**) shows polar plots of the force production and angle of attack of each trajectory.

The forces are defined in the x*-y* directions, which are represented in (**C**). Angle of attack is measured azimuthally. (**C**) shows the optimal trajectories for side force production in two reference frames. The left shows the trajectories with major axes aligned and resultant forces plotted. The right shows the trajectories rotated such that their resultant forces are all in the x* direction. The angle of attack is plotted for the median amputated fin solution. The fin progresses counterclockwise, and the angle of attack of the fin stem is plotted every 15 azimuthal degrees.

**Sensitivity Analysis**

Examining the sensitivity reveals that the optimization landscape for side force is more challenging to navigate than for thrust. We perform principal component analysis (PCA) on the final covariance matrix calculated by the CMA-ES algorithm, scaled to unit variance, to compute sensitivity. This standardized covariance matrix represents the optimal basin-- a hyperellipsoid, whose axis directions and relative lengths are given by PCA. When comparing the normalized scree plots for all thrust trials versus all side force trials (Figure 4) one can see that the side force trials experience a faster dropoff. The faster dropoff means that the side force optimal basin has a more elongated shape than the thrust optimal basin, where several axes have roughly equal magnitude. A more elongated optimal basin is more difficult to traverse since more of its axes have a low eigenvalue and are therefore more sensitive.

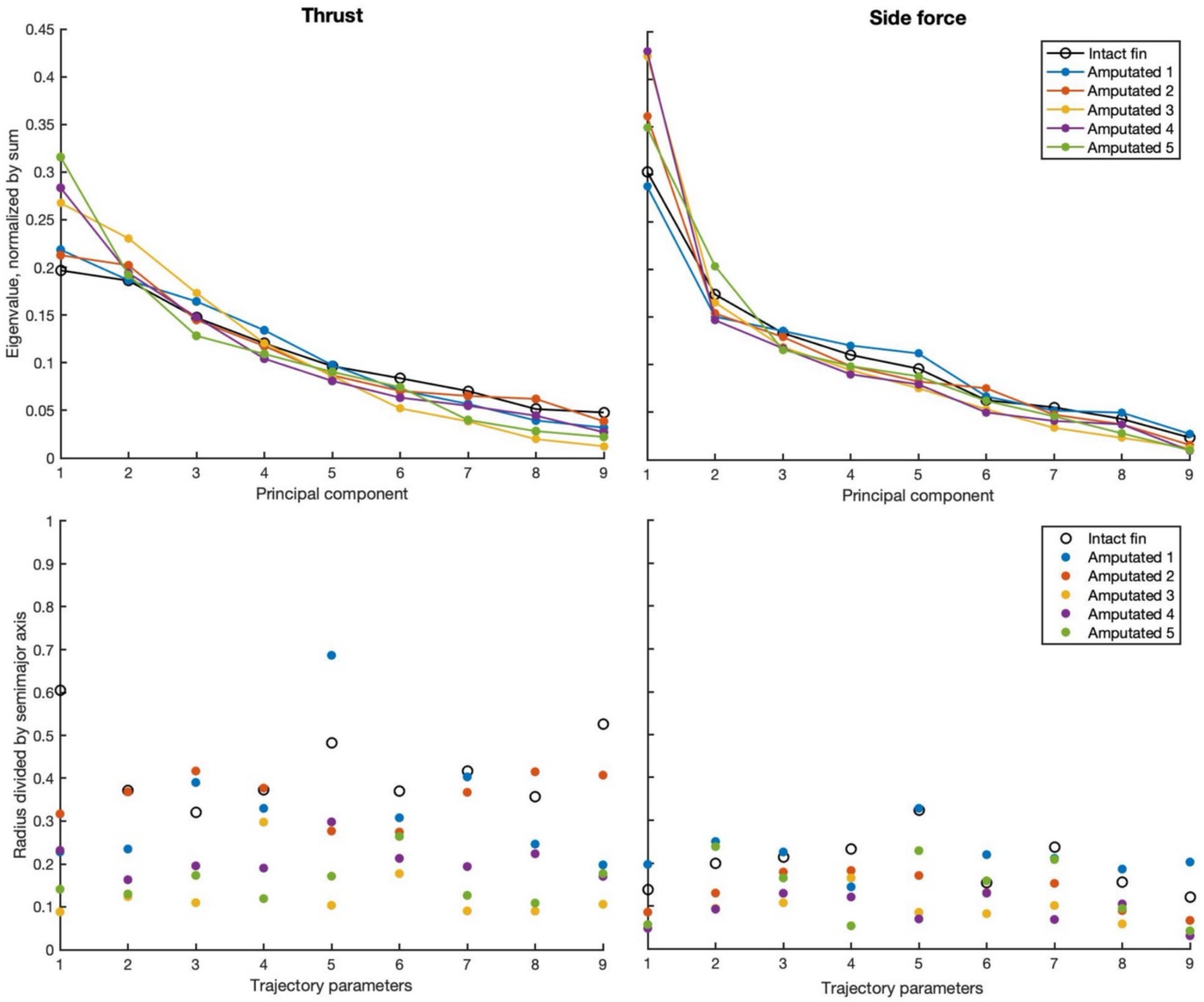

**Fig. 4. Sensitivity analysis via scree plots and hyperellipsoid radii.** The left and right columns are for thrust and side force, respectively. The first row displays the scree plots derived from PCA. The second row plots the normalized radius of each trajectory parameter, where a larger normalized radius corresponds to a lower sensitivity. In order from 1 to 9, the trajectory parameters are: stroke angle, thickness angle, rotation angle, rotation phase, speed code, speed-up value, rotation acceleration, camber, and frequency.

By finding the radius from the origin to the hyperellipsoid's surface along the direction corresponding to a single trajectory parameter, we can translate PCA-based sensitivity into a value for each real-world trajectory parameter. We normalize the calculated radius by the largest eigenvalue defining that hyperellipsoid. The larger the normalized radius, the less sensitive the trajectory parameter is to perturbation. In Figure 4, we compare the normalized radii for the thrust and side force optimizations. The normalized radii of the side force optimization are on average lower, meaning that the sensitivity to perturbation is higher and therefore the optimization landscape more challenging to traverse. However, no other clear trends appear among the data, suggesting that all trajectory parameters are roughly equally involved in the optimization due to their approximately equal sensitivities.

## DISCUSSION

When optimizing for thrust, a full recovery from breakage is seen for both fitness and force production by altering frequency, amplitude, and angle of attack. Because they recover fully in all trials, the robotically actuated trajectories outperform the biological average. In fish, it depends on the species whether accommodation of approximately 50% damage is possible. By comparing the force traces of the intact and amputated fins, we see a qualitative return of force production to its predamaged baseline. Both the amplitude and frequency are increased to maintain force production. This increase matches the theoretical results of (*36*) but does not perfectly align with what is seen in vivo. Although fish do increase their tail beat amplitude upon breakage, they do not display a uniform frequency alteration after breakage. This nonuniformity may suggest that other evolutionary pressures not relevant to the robotic use case affect fishes' frequency adaptations. Furthermore, the angle of attack is significantly altered in the amputated fins as compared to the intact fin. The maximum angle of attack increases, as would be expected to increase force production, and a large phase shift is seen in the robotic optimization. Neither of these angle of attack modifications have been reported in the biological literature as far as the authors are aware.

Some differences between fish and machine-learned adaptations can be explained from the perspective of limited caudal fin actuation. The learned trajectory is easily able to compensate for 50% damage, while some fish may or may not be able to, due to the larger actuation range of the spherical parallel manipulator (SPM). The SPM is capable of actuating inside a much larger angular cone than most fish caudal fins can. This capability, combined with the result that the amputated fins converged to a stroke amplitude near the maximum mechanical constraint of the SPM, implies that the machine-learned adaptation shows improved compensation for damage because it can attain a larger range of motion.

When optimizing for side force, we encounter a more complex optimization landscape, leading to worse recoveries in both fitness and force production after breakage. In fact, one amputated trial was unable to recover to within 2% of the desired force production at all. These results imply that the optimization has a more difficult time accommodating for the breakage when optimizing for side force as compared to thrust, likely due to the optimization function landscape. Because the side force optimal basins are more elongated than the thrust optimal basins, more of the principal component axes and therefore more of the trajectory parameters display high sensitivity. This high sensitivity results in more difficulty accommodating for breakage.

A higher sensitivity, and therefore more difficult optimization, is further reflected in the effects seen on amplitude, frequency, and angle of attack. The amputated trials show a nearly even split on whether a large increase in amplitude or an approximately equal amplitude is optimal to recover from breakage. This split is consistent with the biological literature, which similarly does not reach a consensus.

The effect on frequency is indistinguishable; amputated trials displayed higher, lower, and equal frequencies to the intact fin. This spread contrasts with the biological literature, where most insects increase frequency in response to damage. However, (*10*) noted that increasing wingbeat frequency is likely motivated by behaviors critical to predator evasion. Without the observed increase in frequency, the fruit flies studied would need to utilize their maximum body-constrained wingbeat amplitude during nominal flight conditions, leaving no possibility for increased-speed maneuvers. Increasing frequency allowed the flies to gain a margin of safety between amplitude of nominal operation and maximum amplitude. This suggests that increasing frequency is an adaptation driven by evolutionary pressures that are not relevant to the robotic use case, and therefore is not optimal when only considering efficiency.

Finally, the optimal angle of attack trace was found to nest for all cases, when viewed in the resultant-based reference frame. This includes the intact case, which did not show a large phase shift as was evident in the thrust trials. The nesting of angle of attack optima is quite interesting, as it suggests that the directionality of force production in the x-y plane can be controlled by adding or subtracting angle of attack throughout the cycle.

This work has shown that breakage can easily be accommodated on a flapping propulsor creating thrust by increasing amplitude and frequency as well as significantly altering the magnitude and phase of angle of attack. Accommodating for breakage on a flapping propulsor creating side force, however, is more complex and therefore deserves future focus. The optimal amplitude and frequency alterations to cope with breakage whilst producing side force are still unclear, but one conclusion we may draw to motivate future work is from the angle of attack. All trials, whether intact or amputated, nested with similarly shaped optimal angle of attack traces when in a resultant-based reference frame. To clarify optimal amplitude and frequency strategies, we intend to fix the optimal angle of attack trace and optimize over the remaining trajectory parameters. Minimizing optimization parameters will allow faster convergence of CMA-ES, and therefore the ability to generate a larger number of trials in a time-efficient manner. A subsequent statistical analysis of a larger dataset may clarify how amplitude and frequency should be altered to accommodate for breakage on a flapping propulsor creating side force.

## SUPPLEMENTARY TABLES

| Organism | Force type | Propulsive area removed (%) | Amplitude change | Frequency change | Force change |
|---|---|---|---|---|---|
| *Scardineus erythrophthalmus* (*12*) | Thrust | 100 | Increase | Increase | No change |
| *Merlangius merlangus* (*15*) | Thrust | 100 | Increase | Increase | |
| *Oncorhynchus nerka* (*14*) | Thrust | approx. 33 | No change | No change | No change |
| *Oncorhynchus nerka* (*14*) | Thrust | approx. 66 | No change | No change | Decrease |
| *Etheostoma edwini* (*16*) | Thrust | approx. 50 | | | No change |
| *Gambusia holbrooki* (*13*) | Thrust | 1-30 | | | Gaussian |
| *Spinibarbus sinensis* (*2*) | Thrust | 100 | Increase | No change | Decrease |
| *Cyprinius carpio* (*2*) | Thrust | 100 | Increase | Increase | Decrease |
| *Carassius auratus* (*2*) | Thrust | 100 | Increase | Increase | Decrease |
| *Spinibarbus sinensis* (*2*) | Thrust | 59 | Increase | No change | Decrease |
| *Cyprinius carpio* (*2*) | Thrust | 53 | No change | No change | Decrease |
| *Carassius auratus* (*2*) | Thrust | 76 | No change | Increase | No change |
| *Mylopharyngodon piceus* (*11*) | Thrust | approx. 50 | | No change | No change |
| *Mylopharyngodon piceus* (*11*) | Thrust | 100 | | Decrease | Decrease |
| *Pontia occidentalis* (*9*) | Side force | approx. 15-20 | | Increase | No change |
| *Bombus terrestris* (*3*) | Side force | 10 | No change | Increase | |
| *Limantria dispar* (*7*) | Side force | 44 | | Increase | Decrease |
| *Pieris rapae* (*7*) | Side force | 49 | | No change | Decrease |
| *Bombus impatiens* (*4*) | Side force | approx. 25 | No change | Decrease | Decrease |
| *Manduca sexta* (*8*) | Side force | 10-36 | No change | Increase | No change |
| *Apis mellifera* (*5*) | Side force | 3-37 | Increase | Increase | Decrease |
| *Bombus impatiens* (*6*) | Side force | 11.2 - 23.4 | | Increase | Decrease |
| *Drosophila hydei* (*10*) | Side force | 10-50 | Increase | Increase | No change |

**Table S1. Adaptations to damage of natural swimmers and flyers.** Tabular summary of previous work on natural swimmers' and flyers' adaptations to catastrophic propulsor damage. Boxes left blank were not investigated by the cited authors. "Force type" denotes direction of force produced by the organism in the nomenclature of the robotic system. Force change is listed as an abstraction of change in critical speed, hovering ability, or other parameter revealing force trends.

| | **Intact Fin** | **Amp. 1** | **Amp. 2** | **Amp. 3** | **Amp. 4** | **Amp. 5** | **Intact Fin** | **Amp. 1** | **Amp. 2** | **Amp. 3** | **Amp. 4** | **Amp. 5** |
|---|---|---|---|---|---|---|---|---|---|---|---|---|
| **Force** | Thrust | Thrust | Thrust | Thrust | Thrust | Thrust | SF | SF | SF | SF | SF | SF |
| **Type** | Ellipse | Ellipse | Ellipse | Ellipse | Ellipse | Ellipse | Ellipse | Ellipse | Ellipse | Ellipse | Ellipse | Ellipse |
| **Stroke angle** | 24.8° | 31.7° | 32.0° | 31.5° | 31.5° | 30.0° | 15.2° | 32.1° | 31.3° | 29.6° | 18.7° | 16.8° |
| **Thickness angle** | 12.9° | 15.2° | 14.8° | 15.2° | 13.6° | 13.4° | 8.1° | 15.0° | 15.2° | 13.1° | 13.7° | 11.6° |
| **Rotation angle** | 26.7° | 51.8° | 70.0° | -68.2° | -63.4° | -70.0° | -70.0° | -57.0° | -70.0° | -70.0° | -70.0° | -41.2° |
| **Rotation phase [rad]** | 5.3 | 2.3 | 2.7 | 4.7 | 6.1 | 5.8 | 2.9 | 2.9 | 4.2 | 4.3 | 4.8 | 4.4 |
| **Speed code** | 0 | 1 | 2 | 2 | 2 | 1 | 3 | 4 | 3 | 2 | 3 | 4 |
| **Speed-up value** | 1.2 | 1.2 | 1.2 | 1.2 | 1.2 | 1.2 | 1.1 | 1.2 | 1.2 | 1.1 | 1.1 | 1.1 |
| **Rotation accel.** | 0.6 | 0.8 | 0.5 | 0.5 | 0.9 | 0.5 | 0.2 | 0.4 | 0.4 | 0.4 | 0.3 | 0.3 |
| **Camber** | 0.6 | 0.9 | 0.7 | 1.0 | 0.7 | 0.6 | 0.3 | 0.4 | 0.8 | 0.7 | 0.6 | 0.4 |
| **Frequency [Hz]** | 0.70 | 0.78 | 0.70 | 0.72 | 0.73 | 0.79 | 0.86 | 0.90 | 0.70 | 0.75 | 0.86 | 0.88 |
| **Closeness to setpoint force** | 0.0029 | 0.0130 | 0.0067 | 0.0032 | 0.0085 | 0.0016 | -0.0062 | -0.1618 | -0.0091 | -0.0027 | -0.0189 | 0.0198 |
| **Fitness** | 0.1157 | 0.1088 | 0.0920 | 0.1049 | 0.0879 | 0.0810 | 0.0087 | 0.1633 | 0.0854 | 0.0933 | 0.1059 | 0.0408 |

**Table S2. Converged optimal trajectory parameters and fitness values.** SF abbreviates side force, amp. abbreviates amputated and accel. abbreviates acceleration.

| | | | Mode 1 | Mode 2 | Mode 3 | Mode 4 | Mode 5 |
|---|---|---|---|---|---|---|---|
| **Thrust, $F_z$ (N)** | | | | | | | |
| | **Intact Fin** | **Amplitude** | 0.426 | 0.398 | 0.021 | 0.011 | 0.011 |
| | | **Phase** | ±159.56° | ±48.89° | ±177.34° | ±120.36° | ±161.67° |
| | **Amputated 1** | **Amplitude** | 0.635 | 0.621 | 0.218 | 0.056 | 0.047 |
| | | **Phase** | ±103.49° | ±76.55° | ±20.81° | ±145.21° | ±142.06° |
| | **Amputated 2** | **Amplitude** | 0.516 | 0.369 | 0.184 | 0.062 | 0.056 |
| | | **Phase** | ±23.48° | ±125.74° | ±9.98° | ±93.37° | ±5.65° |
| | **Amputated 3** | **Amplitude** | 0.529 | 0.247 | 0.212 | 0.151 | 0.125 |
| | | **Phase** | ±62.16° | ±92.58° | ±153.66° | ±123.06° | ±28.40° |
| | **Amputated 4** | **Amplitude** | 0.489 | 0.173 | 0.117 | 0.046 | 0.040 |
| | | **Phase** | ±38.45° | ±127.65° | ±15.77° | ±21.72° | ±164.29° |
| | **Amputated 5** | **Amplitude** | 0.623 | 0.413 | 0.086 | 0.059 | 0.046 |
| | | **Phase** | ±15.93° | ±141.96° | ±58.95° | ±21.01° | ±163.90° |
| **Thrust, AOA (°)** | | | | | | | |
| | **Intact Fin** | **Amplitude** | 12.470 | 1.307 | 1.071 | 0.104 | 0.096 |
| | | **Phase** | ±157.52° | ±94.07° | ±29.12° | ±95.22° | ±177.38° |
| | **Amputated 1** | **Amplitude** | 24.078 | 2.421 | 1.730 | 0.894 | 0.334 |
| | | **Phase** | ±33.73° | ±130.23° | ±94.90° | ±144.40° | ±54.98° |
| | **Amputated 2** | **Amplitude** | 34.141 | 2.062 | 1.694 | 0.760 | 0.344 |
| | | **Phase** | ±51.12° | ±147.45° | ±147.42° | ±164.05° | ±137.80° |
| | **Amputated 3** | **Amplitude** | 31.820 | 2.477 | 1.310 | 0.872 | 0.466 |
| | | **Phase** | ±26.88° | ±127.26° | ±142.02° | ±1.54° | ±130.22° |
| | **Amputated 4** | **Amplitude** | 32.261 | 1.675 | 1.097 | 0.665 | 0.357 |
| | | **Phase** | ±64.60° | ±124.38° | ±90.83° | ±158.41° | ±163.57° |
| | **Amputated 5** | **Amplitude** | 34.161 | 1.791 | 1.493 | 0.831 | 0.528 |
| | | **Phase** | ±58.25° | ±36.30° | ±126.34° | ±165.65° | ±154.59° |
| **SF, $F_{x^*}$ (N)** | | | | | | | |
| | **Intact Fin** | **Amplitude** | 0.398 | 0.330 | 0.261 | 0.055 | 0.037 |
| | | **Phase** | ±18.06° | ±9.30° | ±74.55° | ±7.51° | ±127.28° |
| | **Amputated 1** | **Amplitude** | 0.327 | 0.205 | 0.120 | 0.085 | 0.062 |
| | | **Phase** | ±26.50° | ±165.29° | ±15.78° | ±3.84° | ±143.56° |
| | **Amputated 2** | **Amplitude** | 0.623 | 0.320 | 0.107 | 0.091 | 0.057 |
| | | **Phase** | ±77.79° | ±23.67° | ±118.16° | ±89.96° | ±30.22° |
| | **Amputated 3** | **Amplitude** | 0.632 | 0.423 | 0.124 | 0.094 | 0.080 |
| | | **Phase** | ±43.22° | ±122.70° | ±31.45° | ±84.18° | ±131.92° |
| | **Amputated 4** | **Amplitude** | 1.026 | 0.319 | 0.081 | 0.073 | 0.029 |
| | | **Phase** | ±59.48° | ±140.94° | ±156.39° | ±8.42° | ±95.61° |
| | **Amputated 5** | **Amplitude** | 0.507 | 0.236 | 0.108 | 0.049 | 0.020 |
| | | **Phase** | ±129.22° | ±52.89° | ±32.76° | ±119.62° | ±85.85° |
| **SF, $F_{y^*}$ (N)** | | | | | | | |
| | **Intact Fin** | **Amplitude** | 0.311 | 0.161 | 0.136 | 0.043 | 0.041 |
| | | **Phase** | ±13.70° | ±156.69° | ±172.83° | ±92.64° | ±17.64° |
| | **Amputated 1** | **Amplitude** | 0.344 | 0.252 | 0.166 | 0.074 | 0.055 |
| | | **Phase** | ±7.73° | ±40.19° | ±115.02° | ±119.67° | ±50.90° |
| | **Amputated 2** | **Amplitude** | 0.971 | 0.261 | 0.144 | 0.080 | 0.068 |
| | | **Phase** | ±27.17° | ±9.29° | ±144.88° | ±63.86° | ±104.22° |
| | **Amputated 3** | **Amplitude** | 1.099 | 0.339 | 0.110 | 0.106 | 0.096 |
| | | **Phase** | ±64.39° | ±80.68° | ±44.30° | ±134.95° | ±97.31° |
| | **Amputated 4** | **Amplitude** | 1.059 | 0.270 | 0.079 | 0.054 | 0.050 |
| | | **Phase** | ±72.67° | ±50.24° | ±145.82° | ±131.27° | ±12.61° |
| | **Amputated 5** | **Amplitude** | 0.454 | 0.189 | 0.072 | 0.059 | 0.010 |
| | | **Phase** | ±121.11° | ±159.35° | ±156.15° | ±28.10° | ±70.54° |
| **SF, AOA (°)** | | | | | | | |
| | **Intact Fin** | **Amplitude** | 34.840 | 1.137 | 0.443 | 0.246 | 0.174 |
| | | **Phase** | ±11.87° | ±174.42° | ±6.46° | ±76.56° | ±147.26° |
| | **Amputated 1** | **Amplitude** | 27.753 | 2.173 | 1.834 | 0.487 | 0.200 |
| | | **Phase** | ±7.86° | ±97.60° | ±6.82° | ±89.40° | ±173.98° |
| | **Amputated 2** | **Amplitude** | 33.648 | 1.865 | 1.722 | 1.028 | 0.433 |
| | | **Phase** | ±42.04° | ±163.96° | ±43.42° | ±13.47° | ±66.99° |
| | **Amputated 3** | **Amplitude** | 33.566 | 1.211 | 1.082 | 0.828 | 0.481 |
| | | **Phase** | ±49.85° | ±54.73° | ±56.28° | ±0.25° | ±44.98° |
| | **Amputated 4** | **Amplitude** | 34.395 | 0.990 | 0.724 | 0.479 | 0.221 |
| | | **Phase** | ±60.26° | ±147.83° | ±19.17° | ±14.79° | ±45.92° |
| | **Amputated 5** | **Amplitude** | 19.955 | 1.038 | 0.473 | 0.250 | 0.068 |
| | | **Phase** | ±102.73° | ±157.76° | ±74.68° | ±134.85° | ±52.74° |

**Table S3. Fourier decompositions of optimal force and angle of attack traces.** The first five Fourier modes are reported. SF abbreviates side force. Note that the angle of attack Fourier decompositions are reported in the x-y frame for thrust, whilst for side force they are reported in the x*-y* frame.

## ACKNOWLEDGEMENTS

**Funding:**
Center for Autonomous Systems and Technologies (CAST) at the California Institute of Technology
National Science Foundation Graduate Research Fellowship DGE-1745301 (MLH)
Russell R. Vought Scholarship (MLH)

**Author contributions:**
Conceptualization: MG, MLH
Methodology: MG, MLH
Data Curation: MLH
Formal Analysis: MLH, IS
Investigation: MLH
Visualization: MLH, IS
Funding acquisition: MG, MLH
Resources: MG
Project administration: MG
Supervision: MG, IS
Writing – original draft: MLH
Writing – review & editing: MG, MLH, IS

**Competing interests:**
Authors declare that they have no competing interests.

**Data and materials availability:**
All data are in the main text or the supplementary materials.